\documentclass[sigconf, screen]{acmart}
\AtBeginDocument{%
  }
\usepackage{etoc}
\usepackage{titletoc}
\usepackage{tcolorbox}
\setcopyright{acmlicensed}
\copyrightyear{2018}
\acmYear{2018}
\acmDOI{XXXXXXX.XXXXXXX}
\acmConference[Conference acronym 'XX]{Make sure to enter the correct
  conference title from your rights confirmation email}{June 03--05,
  2018}{Woodstock, NY}
\acmISBN{978-1-4503-XXXX-X/2018/06}

\begin{document}

\title{RAGOCR: Optical Compression of Retrieval-Augmented Text via Visual Representation}

\author{Jiayang Yu}
\affiliation{%
  \institution{Wangxuan Institute of Computer Technology, Peking University}
  \city{Beijing}
  \country{China}
}
\email{yujiayang26@stu.pku.edu.cn}

\author{Jialun Zhong}
\affiliation{%
  \institution{Wangxuan Institute of Computer Technology, Peking University}
  \city{Beijing}
  \country{China}
}
\email{zhongjl@stu.pku.edu.cn}

\author{Lei Zou}
\affiliation{%
  \institution{Wangxuan Institute of Computer Technology, Peking University}
  \city{Beijing}
  \country{China}
}
\email{zoulei@pku.edu.cn}

\renewcommand{\shortauthors}{Trovato et al.}

\begin{abstract}

  Retrieval-Augmented Generation (RAG) has become essential for knowledge-intensive question answering, yet scaling RAG pipelines remains challenging due to the prohibitive computational cost of processing lengthy retrieved contexts. Existing compression approaches face a fundamental trade-off: hard compression methods operate online in a query-aware fashion but achieve only modest compression rates and typically require fine-tuning the generative model, while soft compression methods attain higher ratios but rely on costly offline encoding that is entirely agnostic to the input query. To bridge this gap, we introduce RAGOCR, a novel framework that compresses retrieved documents into compact visual representations conditioned on the input query. To further balance compression rate and information fidelity, we introduce a query-aware dynamic resolution mechanism that adaptively allocates visual granularity based on each document's estimated relevance and complexity: highly relevant passages are rendered at higher resolution to preserve fine-grained details, while peripheral documents are aggressively compressed at lower resolution. Experiments on five QA benchmarks using the MedOmniKB retrieval corpus demonstrate that RAGOCR surpasses naive RAG by over 15\% in accuracy while requiring only one-eighth the number of input tokens, and consistently outperforms both hard and soft compression baselines across varying retrieval depths.
 \end{abstract}




\maketitle

\section{Introduction}

\begin{figure}[t]
  \includegraphics[width=\textwidth, trim=0 0 120 0, clip]{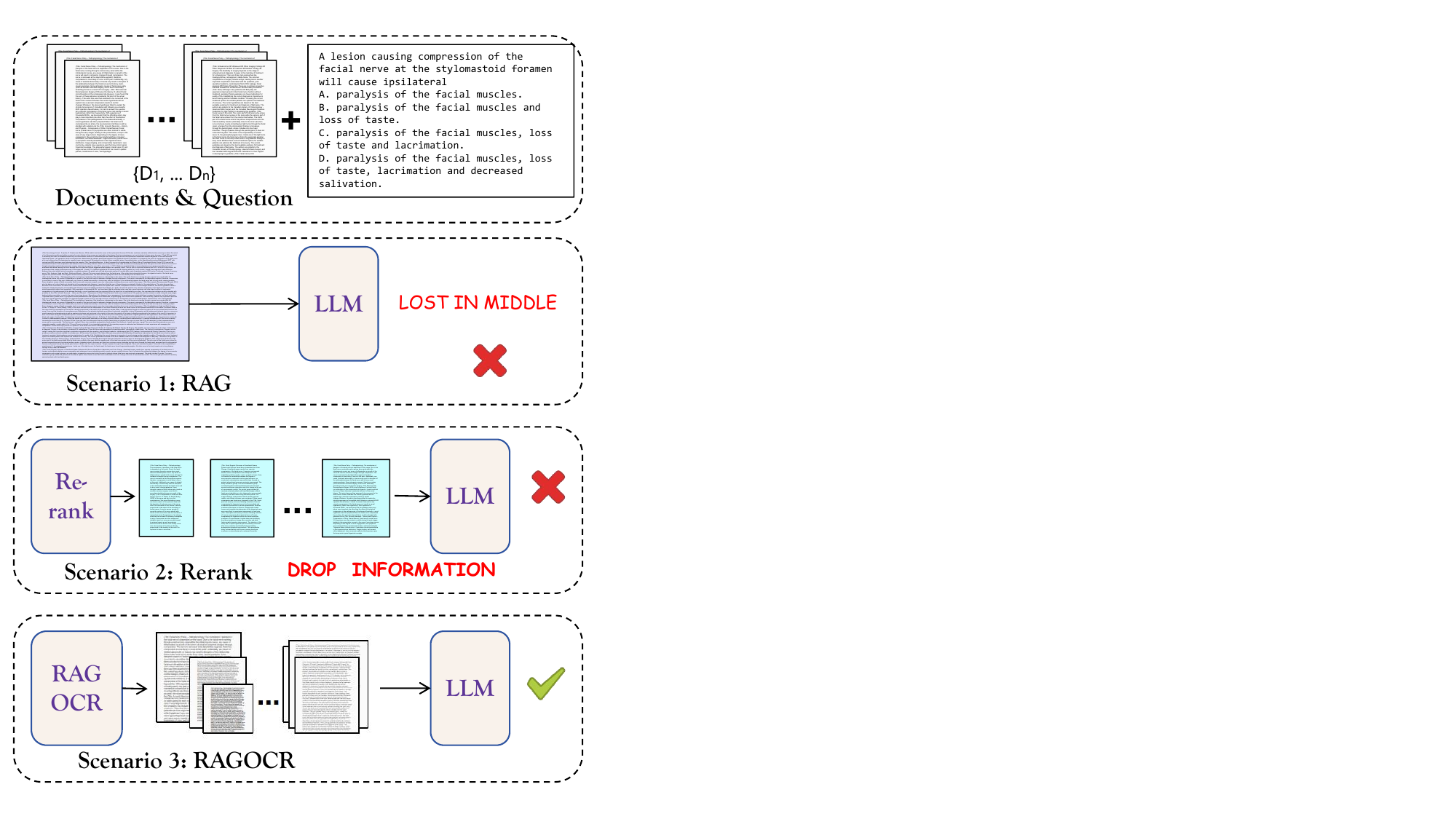}
  \caption{Comparison of three RAG paradigms. (a) RAG concatenates retrieved passages as text, where the model struggles to relevant information buried in the middle of lengthy contexts. (b) Reranking filters documents by relevance scores, reducing noise but risking the loss of supporting evidence in knowledge-intensive scenarios. (c) RAGOCR renders all documents as images and adaptively compresses each, preserving critical details at high resolution while aggressively downscaling less pertinent content, thereby retaining more information within a compact token budget.}
  \Description{Comparison of three RAG paradigms. (a) RAG concatenates retrieved passages as text, where the model struggles to relevant information buried in the middle of lengthy contexts. (b) Reranking filters documents by relevance scores, reducing noise but risking the loss of supporting evidence in knowledge-intensive scenarios. (c) RAGOCR renders all documents as images and adaptively compresses each, preserving critical details at high resolution while aggressively downscaling less pertinent content, thereby retaining more information within a compact token budget.}
  \label{fig:method}
\end{figure}

Retrieval-Augmented Generation (RAG)~\cite{lewis2020retrieval} has emerged as a promising paradigm for enhancing the factual accuracy of Large Language Models (LLMs) in knowledge-intensive environments. By prepending retrieved passages to the input query, LLMs can leverage up-to-date external knowledge without costly retraining, making RAG particularly appealing for tasks such as open-domain question answering, fact verification, and knowledge-grounded dialogue~\cite{guu2020retrieval, wang2025rare}.

However, a fundamental tension exists in current RAG systems: \textbf{more context helps, but longer context hurts}. On the one hand, incorporating a sufficient number of relevant documents substantially improves response quality. On the other hand, as the volume of retrieved content grows, LLMs struggle to attend to critical information buried within lengthy inputs, leading to a significant degradation in generation quality~\cite{du2025context,liu2024lost}. This dilemma has motivated a line of research on context compression, which seeks to distill the retrieved documents into more compact representations before feeding them to the generator. Existing approaches include token- or sentence-level pruning that removes less informative content from the original text~\cite{jiang-etal-2023-llmlingua,jiang-etal-2024-longllmlingua,pan-etal-2024-llmlingua,li2025securitylingua}, and abstractive methods that replace original documents with shorter summaries~\cite{xu2023recomp}. While effective to varying degrees, these methods operate entirely within the text modality, where compression is inherently constrained by the discrete, one-dimensional nature of token sequences.

A recent and striking finding offers a fundamentally different perspective: \textbf{visual tokens carry substantially higher information density than text tokens}. As demonstrated by DeepSeek-OCR~\cite{wei2025deepseek}, a document page can be faithfully decoded from as few as 100 vision tokens with 97\% precision, achieving up to 10$\times$ compression over equivalent text token representations. Current work such as Glyph~\cite{cheng2025glyph} and Vist~\cite{xing2025vision} further confirms that rendering text into images and processing them through vision-language models can substantially reduce token consumption while preserving semantic fidelity. More broadly, operating directly in the pixel space has proven effective for generative modeling as well~\cite{yan2026pixel}, further motivating a visual, pixel-level treatment of textual content. This insight opens up an appealing opportunity for RAG: 
by rendering retrieved documents as images, we can convey richer information within a fixed token budget, enabling models to absorb more knowledge without exceeding their context window limits, or transmit the same amount of information with substantially fewer tokens, significantly reducing inference cost.

However, naively applying this render-based paradigm to RAG introduces a critical challenge: retrieved documents vary drastically in their relevance to the query. Applying a uniform resolution across all rendered images inevitably leads to either wasteful token expenditure on marginally relevant passages or catastrophic information loss on highly pertinent ones. Unlike general long-context compression, the RAG scenario naturally demands a \textbf{query-aware} allocation strategy that dynamically assigns visual resolution based on the semantic relevance between each document and the query.

To address this challenge, we propose \textbf{RAGOCR} (\textbf{R}etrieval-\textbf{A}ugmented \textbf{G}eneration via \textbf{O}ptical \textbf{C}haracter \textbf{R}endering), a framework that dramatically reduces the token cost of retrieved documents through adaptive visual compression. Specifically, RAGOCR operates in two stages: (1) rendering each retrieved document as an image, and (2) compressing each image to a query-appropriate resolution. We employ Group Relative Policy Optimization (GRPO)~\cite{shao2024deepseekmath} to train a lightweight compressor that adaptively assigns a compression ratio to each rendered passage based on its relevance to the query.

Interestingly, we discover that the compression ratios produced by our compressor serve as a surprisingly effective proxy for passage relevance. Since the compressor learns to allocate higher resolution to more query-relevant passages, its output ratios naturally encode a relevance ranking. Experiments show that using these ratios directly as reranking scores outperforms dedicated reranking models such as Qwen3-Reranker-4B~\cite{qwen3embedding}, and combining RAGOCR with it yields further improvements. This finding suggests that optical compression and relevance estimation are two sides of the same coin, pointing toward a unified paradigm where a single model simultaneously compresses and ranks retrieved content.

Our main contributions are summarized as follows:

\begin{itemize}
    \item We introduce RAGOCR, a framework that combines optical rendering with dynamic resolution allocation to compress text-based RAG, achieving a strong balance between token efficiency and answer quality.
    \item We develop a query-aware compressor trained with GRPO that adaptively assigns compression ratios to each rendered passage based on its relevance to the query.
    \item We conduct comprehensive experiments on multiple benchmarks, demonstrating that RAGOCR consistently outperforms existing methods. Moreover, our experiments show that RAGOCR generalizes well across diverse knowledge sources, including free-text passages, knowledge graphs, and visually rich image-based documents.
    \item We reveal that the compression ratios learned by RAGOCR inherently encode passage relevance, enabling the compressor to double as a reranker that surpasses dedicated reranker models, and we discuss the possibility of building unified compression-reranker systems.
\end{itemize}
\section{Related Work}

\subsection{Context Compression for RAG}

Retrieval-augmented generation (RAG)~\cite{lewis2020retrieval} enhances large language models by grounding their responses in externally retrieved evidence, yet the growing volume of retrieved passages introduces substantial computational overhead and noise propagation~\cite{shi2024replug}. Context compression has emerged as a key technique to mitigate these issues, and existing methods can be broadly categorized into hard compression and soft compression.

Hard compression methods operate directly on the surface text by removing or rewriting tokens. At the token level, LongLLMLingua~\cite{jiang2024longllmlingua} and LLMLingua-2~\cite{pan2024llmlingua} leverage perplexity-based scoring to prune low-information tokens from the prompt. At the sentence level, RECOMP~\cite{xu2023recomp} proposes both an extractive variant that selects top-ranked sentences and an abstractive variant that generates compressed summaries. Provence~\cite{chirkova2025provence} formulates context pruning as sequence labeling and unifies it with document reranking in a single lightweight cross-encoder, dynamically adjusting the pruning ratio per query-context pair. EXIT~\cite{hwang2025exit} further improves extractive compression by performing context-aware sentence classification that adapts to query complexity and retrieval quality. FaviComp~\cite{jung2024familiarity} takes an orthogonal approach by generating abstractive compressions through ensemble decoding, lowering the perplexity of the compressed evidence with respect to the target model in a training-free manner.

Soft compression methods, in contrast, map retrieved documents into compact continuous representations consumed directly by the generator LLM. xRAG~\cite{cheng2024xrag} achieves extreme compression by projecting dense retrieval embeddings into the LLM's representation space through a lightweight modality bridge. PISCO~\cite{louis2025pisco} distills document-level knowledge into fixed-length embedding vectors via sequence-level knowledge distillation. OSCAR~\cite{louis2025oscar} bridges online hard and offline soft paradigms by performing query-dependent soft compression at inference time.

Despite their effectiveness, all the above methods operate within the conventional text token space, where compression is inherently constrained by the discrete, one-dimensional nature of token sequences. A fundamentally different direction is to leverage the visual modality as a higher-density compression channel for textual information, which we discuss next.

\subsection{Optical Text Compression}

To compress the long input texts, recent research has explored a novel paradigm that leverages the visual modality as a more information-dense medium for representing textual content.
The core insight is that rendering text into images and processing them through vision encoders can yield substantially fewer tokens than conventional text tokenization, since a single visual token inherently encodes richer information including typography, spatial layout, and contextual semantics.
DeepSeek-OCR~\cite{wei2025deepseek} pioneers the concept of \textit{Contexts Optical Compression}, which maps long textual contexts into 2D optical representations.
Its successor, DeepSeek-OCR 2~\cite{wei2026deepseek}, further introduces a Causal Visual Flow mechanism to improve reading-order recognition for complex layouts.
Glyph~\cite{cheng2025glyph} extends this paradigm to long-context language understanding by rendering ultra-long texts into images and processing them with vision-language models.
It employs an LLM-driven genetic search to automatically explore rendering parameters (\textit{e.g.}, font size, layout, resolution) that maximize compression while preserving downstream accuracy.
Vist~\cite{xing2025vision} proposes a slow-fast compression framework inspired by human selective reading.
The fast path renders distant, low-salience context into images and processes them with a frozen lightweight vision encoder, while the slow path feeds the proximal tokens directly into the LLM for fine-grained reasoning.

While these methods demonstrate the feasibility of optical compression for long-context modeling, their compression strategies are 
either globally uniform or determined solely by positional proximity, without considering the semantic relevance between the compressed content and the downstream task query. In contrast, our approach targets the RAG scenario, where retrieved passages exhibit varying degrees of relevance to the query. We train a compressor, which can adaptively assign different compression 
ratios, achieving improved results.

\subsection{Reinforcement Learning for RAG}

Reinforcement learning (RL) has recently emerged as a powerful post-training paradigm for large language models. GRPO~\cite{shao2024deepseekmath}, eliminates the need for a separately trained critic by estimating advantages from group-sampled responses, substantially simplifying the training pipeline while achieving strong reasoning performance. RL has likewise become an effective post-training paradigm for other generative models, spanning policy optimization~\cite{yan2025entropy}, step-efficient RL fine-tuning~\cite{yan2026less}, and multi-objective reward design~\cite{yanmultitune} for diffusion models.

Applying RL to various stages of the RAG pipeline grows recently. For the retrieval stage, Search-r1~\cite{jin2025search} train LLMs via RL to autonomously interleave multi-turn reasoning with search engine invocations, learning when and what to retrieve without supervised demonstrations. For the generation stage, RAG-RL~\cite{huang2025rag} applies GRPO with rule-based rewards to train a reasoning reader that can effectively distinguish relevant from distractor passages. For the reranking stage, REARANK~\cite{zhang2025rearank} proposes the first reasoning-based listwise reranking agent optimized with RL.

Existing RL-based approaches for RAG focus on optimizing retrieval, generation, or reranking as separate components, leaving context compression untouched by RL. In our work, we adopt GRPO to train a lightweight compressor that learns to allocate visual resolution across rendered passages based on their query relevance.
\section{Method}

\begin{figure*}[t]
  \includegraphics[width=\textwidth, trim=0 80bp 0 0, clip]{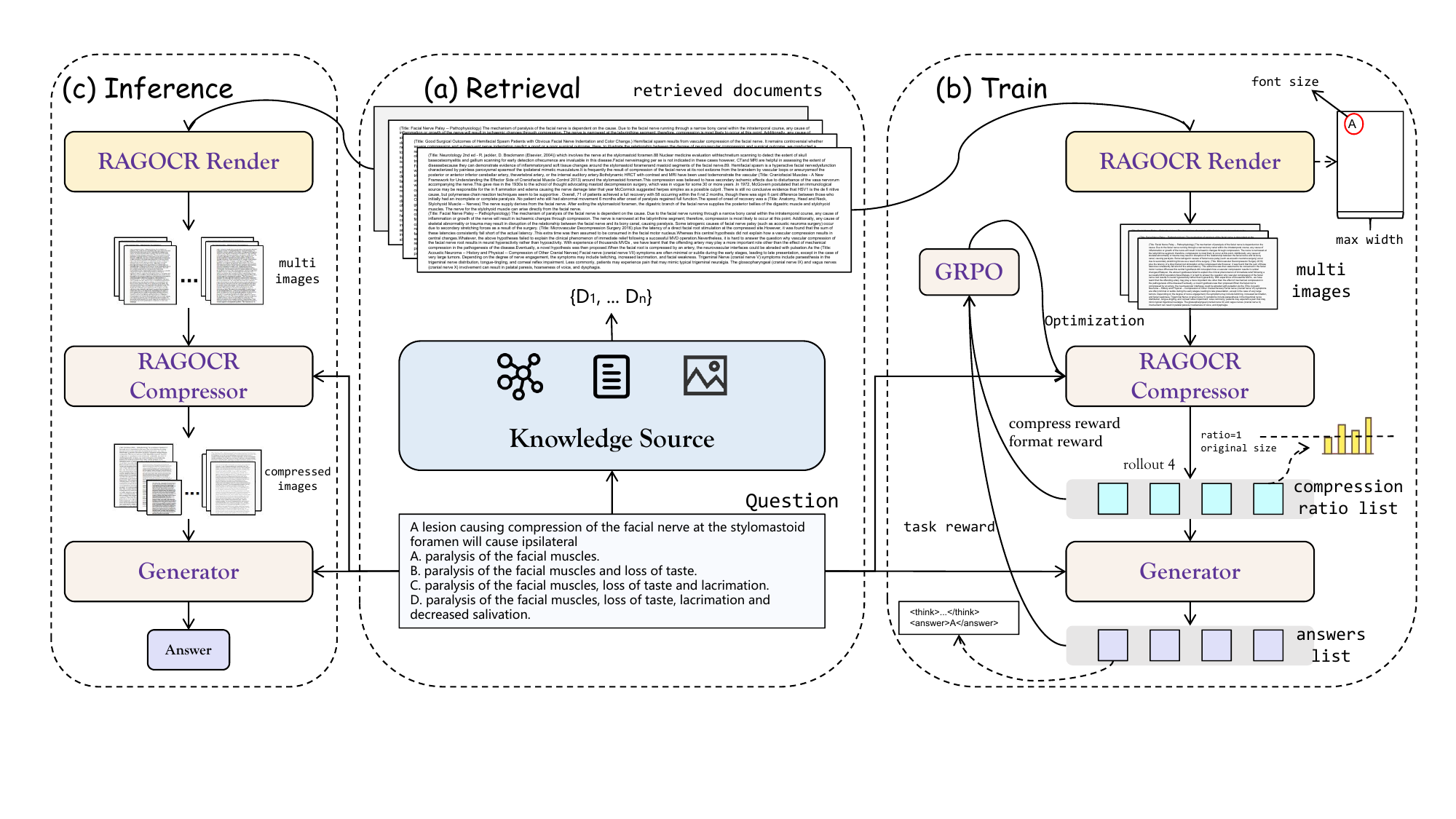}
  \caption{Overview of the RAGOCR framework, which operates effectively across knowledge sources of different modalities. The pipeline consists of three stages.}
  \Description{The RAGOCR framework consisting of three stages: retrieval from a knowledge base, GRPO-based training with task, compression, and format rewards, and efficient two-pass inference with adaptive optical compression.}
  \label{fig:method}
\end{figure*}

\subsection{Preliminaries}

\paragraph{Retrieval-Augmented Generation}
Given a query $q$, a retrieval-augmented generation system first retrieves a set of $N$ documents $\mathcal{D}=\{d_1, d_2, \dots, d_N\}$ from an external knowledge source. A generator model $\mathcal{G}$ then produces an answer 
$\hat{a}$ conditioned on both the query and the retrieved documents:
\begin{equation}
  \hat{a} = \mathcal{G}(q, \mathcal{D}).
\end{equation}
In conventional text RAG, each document $d_i$ is tokenized into a sequence of text tokens and concatenated with the query. As $N$ grows, the total token count increases rapidly, degrading both efficiency and generation quality.

\paragraph{Group Relative Policy Optimization}
GRPO~\cite{shao2024deepseekmath} is a reinforcement learning 
algorithm that optimizes a policy $\pi_\theta$ without requiring a separately trained value model. For each input, GRPO samples a group of $G$ responses $\{y_k\}_{k=1}^G$ from the current policy and evaluates each with a scalar reward $r_k$. The advantage of each sample is computed relative to the group statistics:
\begin{equation}
  A_k = \frac{r_k - \mathrm{mean}(\{r_j\}_{j=1}^{G})}
  {\mathrm{std}(\{r_j\}_{j=1}^{G}) + \epsilon}.
\end{equation}
The policy is updated by maximizing the group-normalized advantage while constraining deviation from a reference policy $\pi_{\mathrm{ref}}$ via a KL penalty weighted by $\beta$.


\subsection{Overview}

As illustrated in Figure~\ref{fig:method}, RAGOCR comprises two stages. In the \textbf{rendering stage} (\S\ref{sec:render}), each retrieved document is converted into an image through text rasterization, transforming the one-dimensional token sequence into a two-dimensional visual representation with inherently higher information density. In the \textbf{optical compression stage} (\S\ref{sec:compress}), a query-aware compressor, trained via GRPO, 
examines all rendered images jointly with the query and predicts a per-image compression factor. Images corresponding to highly relevant documents are assigned lower compression to preserve fine-grained details, while less relevant ones are aggressively down-scaled. The compressed images are then fed to a vision-language generator to produce the final answer. Notably, RAGOCR can be seamlessly integrated into any existing text RAG pipeline, either replacing or complementing a conventional reranker.

\subsection{Rendering Stage} 
\label{sec:render}
Given the set of retrieved documents \(\mathcal{D}=\{d_i\}_{i=1}^N\), we render each document \(d_i\) into an image \(p_i\). This defines a mapping
\begin{equation}
\mathcal{R}: \mathcal{D} \rightarrow \mathcal{P}, \qquad p_i=\mathcal{R}(d_i),
\end{equation}
where \(\mathcal{P}=\{p_i\}_{i=1}^N\) denotes the set of rendered images. The rendering process is parameterized by the font size \(f\), maximum canvas width \(w\), and padding \(m\).
Let \(c(f)\) and \(\ell(f)\) denote the average character width and line height under font size \(f\). The maximum number of characters per line is
\begin{equation}
M = \left\lfloor \frac{w-2m}{c(f)} \right\rfloor.
\end{equation}
We then apply a line-wrapping operator to document \(d_i\),
\begin{equation}
\mathcal{W}_M(d_i)=\{s_{i,1},\dots,s_{i,n_i}\},
\end{equation}
where \(s_{i,j}\) denotes the \(j\)-th rendered line and each line satisfies \(|s_{i,j}| \le M\). The resulting canvas height is determined by the number of wrapped lines:
\begin{equation}
h_i = 2m + n_i \ell(f).
\end{equation}
Finally, the image \(p_i\) is obtained by rasterizing the wrapped lines onto a blank canvas:
\begin{equation}
p_i=\operatorname{Draw}\!\left(w,h_i,\{s_{i,j}\}_{j=1}^{n_i},f\right).
\end{equation}
Applying \(\mathcal{R}\) to all retrieved documents yields the rendered image set \(\mathcal{P}=\{p_i\}_{i=1}^N\).
Since the rendering is deterministic and independent of the query, all images can be precomputed offline, introducing negligible overhead at inference time. As we show in Section~\ref{sec:exp}, even this simple step yields improvements in downstream performance, as the resulting visual tokens encode the same textual content with substantially fewer tokens compared to their text counterparts.

\subsection{Optical Compression Stage}
\label{sec:compress}
The core component of RAGOCR is a query-aware optical compressor, which predicts a per-image compression factor based on the relevance 
of each rendered document to the input query. Intuitively, images that are more informative for answering the question should be assigned lower compression, while less relevant ones can be aggressively down-scaled. We formulate this as a reinforcement learning problem and train a lightweight vision-language model as the compressor policy.

\paragraph{Policy formulation.}

Given a query \(q\) and the rendered image set \(\mathcal{P}=\{p_i\}_{i=1}^{N}\), the compressor policy \(\pi_{\theta}\) generates a structured response \(y\) containing the compression factors for all images:
\begin{equation}
y \sim \pi_{\theta}(\cdot \mid q, \mathcal{P}),
\end{equation}
where from $y$ we parse the compression factors $\{c_i\}_{i=1}^{N}$ with $c_i \ge 1$. Following standard image resizing, the spatial resolution of $p_i$ is reduced by a factor of $1/\sqrt{c_i}$ along each dimension, so that the number of visual tokens decreases proportionally to $c_i$. In our implementation, we adopt Qwen3-VL-2B as the compressor backbone.

\paragraph{GRPO training.}

For each input, we sample a group of responses \(\{y_k\}_{k=1}^{G}\) from the current policy and evaluate each sample with a scalar reward \(r_k\). GRPO optimizes the policy by comparing each sample against the group-relative baseline. The advantage of sample \(y_k\) is computed as
\begin{equation}
\hat{A}_k = \frac{r_k-\mathrm{mean}(\{r_j\}_{j=1}^{G})}{\mathrm{std}(\{r_j\}_{j=1}^{G})+\delta}.
\end{equation}
The policy is then updated by maximizing the normalized group advantage while constraining the deviation from the reference policy \(\pi_{\mathrm{ref}}\):
\begin{equation}
\mathcal{L}_{\mathrm{GRPO}}
= -\,\mathbb{E}\!\left[
  \frac{1}{G}\sum_{k=1}^{G}\frac{1}{|y_k|}\sum_{t=1}^{|y_k|}
  \!\Big(
    \min\!\big(
      \rho_{k,t}\,\hat{A}_k,\;
      \bar{\rho}_{k,t}\,\hat{A}_k
    \big)
    - \beta\, D_{k,t}^{\mathrm{KL}}
  \Big)
\right],
\end{equation}
where $\rho_{k,t} = \pi_{\theta}(y_{k,t}\mid q, y_{k,<t}, \mathcal{P}) \,/\, \pi_{\theta_{\mathrm{old}}}(y_{k,t}\mid q, y_{k,<t}, \mathcal{P})$ is the per-token importance ratio, $\bar{\rho}_{k,t}=\mathrm{clip}(\rho_{k,t},\,1{-}\epsilon,\, 1{+}\epsilon)$, $D_{k,t}^{\mathrm{KL}}$ is the per-token KL divergence between $\pi_{\theta}$ and $\pi_{\mathrm{ref}}$, and $\beta$ controls the KL penalty strength. In practice, we use GRPO because it is well suited for structured generation such as RAGOCR

\paragraph{Compression reward.}
The reward is designed to encourage three properties simultaneously: \emph{task correctness}, \emph{compression efficiency}, and \emph{output format validity}. Given a sampled response \(y\), we first parse the predicted compression factors \(\{c_i\}_{i=1}^{N}\), compress the corresponding images, and feed the compressed images into a fixed generator model to answer the downstream question. The overall reward is defined as
\begin{equation}
r = r_{\mathrm{task}} + \lambda \, r_{\mathrm{comp}} \cdot \mathbb{I}(\text{inject}) + \alpha \, r_{\mathrm{format}},
\end{equation}
where \(\lambda\) is the compression reward weight, \(\alpha\) is the format reward weight, and \(\mathbb{I}(\text{inject})\) indicates whether the compression reward is enabled at the current training step. In our implementation, the compression term is injected only at sparse intervals during training rather than at every update, which helps stabilize early-stage learning.

The task reward measures whether the answer produced by the fixed generator is correct:
\begin{equation}
r_{\mathrm{task}}=
\begin{cases}
1, & \text{if } \hat{a}=a^{\star},\\
0, & \text{otherwise},
\end{cases}
\end{equation}
where \(\hat{a}\) is the generator prediction from the compressed images and \(a^{\star}\) is the ground-truth answer. This reward ensures that the compressor learns to preserve task-relevant information.

The compression reward encourages more aggressive compression, but it is activated only when the downstream answer is correct:
\begin{equation}
r_{\mathrm{comp}}=
\begin{cases}
\frac{1}{N}\sum_{i=1}^{N}\log c_i, & \text{if } r_{\mathrm{task}}=1,\\
0, & \text{otherwise}.
\end{cases}
\end{equation}
This design prevents the model from trivially maximizing compression at the expense of task performance. The logarithm further avoids excessively favoring extremely large compression factors.

Finally, we include a format reward to ensure that the model outputs a valid compression template:
\begin{equation}
r_{\mathrm{format}}=
\begin{cases}
1, & \text{if the output contains a valid format} \\
0, & \text{otherwise}.
\end{cases}
\end{equation}
If the output does not follow the required format, the sample receives zero overall reward and is not forwarded to the downstream generator. This simple rule substantially improves the structural reliability of the compressor outputs.

Overall, the proposed reward encourages the compressor to allocate higher visual budget to informative images and lower budget to less useful ones, while maintaining the downstream QA accuracy under constrained visual tokens. The use of a composite objective that balances multiple competing terms echoes multi-objective optimization strategies explored in other generative settings~\cite{yanmultitune}.

\subsection{Inference Stage}

At inference time, given a query $q$ and its retrieved documents $\mathcal{D}$, RAGOCR first renders each document into an image $p_i = \mathcal{R}(d_i)$. The compressor $\pi_\theta$ then takes the query and all rendered images as input and generates the compression factors $\{c_i\}_{i=1}^{N}$ in a single forward pass. Each image $p_i$ is resized by a factor of $1/\sqrt{c_i}$ along each spatial dimension, reducing its visual token count proportionally to $c_i$.
The compressed image set, together with the query, is then fed to the generator $\mathcal{G}$ to produce the final answer. The entire pipeline requires only two lightweight forward passes, the first one through the compressor and the second one through the generator, with no iterative retrieval or multi-round interaction, making RAGOCR efficient and straightforward to integrate into existing RAG workflows.

\paragraph{Compressor as reranker.} Since the compressor learns to assign lower compression ratios to more query-relevant passages, the inverse of the compression factor $s_i = 1/c_i$ naturally serves as a relevance score. Ranking documents by $s_i$ enables the compressor to function as a reranker without any additional module or training objective. As we show in Section~\ref{sec:exp}, this emergent ranking signal -- arising from the compressor jointly observing all passages in a single forward pass -- outperforms dedicated pointwise rerankers and provides complementary gains when combined with them.
\section{Experiments}
\label{sec:exp}

\subsection{Experimental Setup}
\paragraph{Datasets.}

We select five knowledge-intensive QA benchmarks that are widely adopted for evaluating retrieval-augmented LLMs. MedQA~\cite{jin2021disease} contains 12,723 four-option multiple-choice questions collected from the United States Medical Licensing Examination (USMLE), requiring multi-hop clinical reasoning. MedMCQA~\cite{pal2022medmcqa} comprises more than 194k multiple-choice questions from the Indian medical entrance examinations (AIIMS and NEET-PG), covering 2,400 healthcare topics across 21 subjects. PubMedQA~\cite{jin2019pubmedqa} is a biomedical research QA dataset derived from PubMed abstracts, where each question is answered with yes, no, or maybe. BioASQ~\cite{tsatsaronis2015overview} is a biomedical semantic QA benchmark from which we use the yes/no question subset. MMLU-Med~\cite{hendryckstest2021,hendrycks2021ethics} consists of nine medicine-related subjects from the Massive Multitask Language Understanding benchmark, including clinical knowledge, professional medicine, anatomy, and medical genetics. These benchmarks collectively span diverse question formats and reasoning complexities, providing a comprehensive testbed for evaluating context compression under knowledge-intensive conditions.

We adopt MedOmniKB~\cite{chen2025towards} as the unified retrieval corpus. MedOmniKB is a multi-genre, multi-structured knowledge base that aggregates five heterogeneous sources: textbooks, clinical guidelines, research articles, encyclopedia entries, and structured knowledge graphs. This diversity makes it particularly suited for our setting: it provides documents of varying lengths, structures, and relevance levels per query, creating a challenging and realistic scenario for evaluating adaptive compression. For each query, we retrieve from all five sources with top-$k{=}3$ per source, yielding up to 15 candidate documents that mix free-text documents with graph-derived structured content.

We retain single-answer multiple-choice questions across all five benchmarks. We use the training splits of MedQA, MedMCQA, PubMedQA, and BioASQ to construct the unified training set. For each training and test instance, we pass the question into the MedOmniKB retrieval framework to obtain the corresponding candidate documents, retrieving top-$k{=}3$ from each of the five sources.

\paragraph{Implementation Details}

All experiments are conducted on eight NVIDIA H20 GPUs. We adopt Qwen3-VL-2B-Thinking~\cite{qwen3technicalreport, Qwen2.5-VL, Qwen2VL, Qwen-VL} as the compressor backbone and optimize it with GRPO. During training, four GPUs are allocated for policy optimization and the remaining four are used to host the frozen generator for reward computation. We train the compressor with a group size of $G{=}$4 and a learning rate of 5e-6 using the AdamW optimizer with a cosine schedule. At inference, the compressor processes all 15 retrieved images jointly with the query in a single forward pass and outputs the compression factors, after which the resized images are fed to the generator for answer production.

\paragraph{Metrics.}
We prompt the generator to produce structured responses containing the selected option, and report accuracy as the evaluation metric across all five benchmarks.

\begin{table*}[t]
\caption{Performance comparison on five knowledge-intensive QA benchmarks. All methods except OSCAR share the same generator. RAGOCR achieves the best results across similar size of generators.}
\label{table:performance of RAGOCR}
\renewcommand{\arraystretch}{0.7}
\centering
\resizebox{0.9\textwidth}{!}{
    \begin{tabular}{c|cccccc}
    \toprule
    \textbf{Method} & \textbf{BioASQ} & \textbf{MedMCQA} & \textbf{MedQA} & \textbf{MMLU} & \textbf{PubMedQA} & \textbf{average} \\ 
    \midrule
    Naive RAG~\cite{lewis2020retrieval} & 61.76 & 40.75 & 36.14 & 41.32 & 56.00 & 44.63  \\ 
    Visual RAG & 64.45 & 46.00 & 60.33 & 64.19 & 49.20 & 57.16  \\ 
    Static OCR compress & 61.13 & 44.58 & 58.99 & 62.53 & 44.00 & 55.02  \\ 
    Qwen3-Reranker-4B~\cite{qwen3embedding} & 68.41 & 48.25 & 46.50 & 59.41 & 57.60 & 54.52  \\ 
    Qwen3-VL-Reranker-2B~\cite{qwen3vlembedding} & 70.33 & 50.00 & 64.02 & 64.37 & 52.80 & 60.49  \\ 
    OSCAR-7B~\cite{louis2025oscar} & 77.62 & 35.75 & 34.89 & 50.37 & 31.80 & 45.17  \\ 
    OSCAR-24B~\cite{louis2025oscar} & 80.31 & 51.00 & 60.22 & 67.71 & 51.20 & 61.93  \\ 
    LongLLMLingua~\cite{jiang-etal-2024-longllmlingua} & 71.61 & 50.17 & 43.05 & 54.82 & 58.80 & 53.70  \\ 
    Provence~\cite{chirkova2025provence} & 63.04 & 52.58 & 60.41 & 64.83 & 47.20 & 58.53 \\ 
    Exit~\cite{hwang2025exit} & 77.37 & 51.42 & 39.43 & 54.27 & 56.00 & 53.57  \\
    FaviComp~\cite{jung2024familiarity} & 64.71 & 46.75 & 52.32 & 55.00 & 51.80 & 53.49  \\
    \textbf{RAGOCR} & 71.36 & 45.75 & 69.21 & 68.87 & 58.00 & 62.51  \\
    \midrule
    \end{tabular}
}
\end{table*}

\subsection{Evaluation Details}

To comprehensively evaluate RAGOCR, we compare it against a diverse set of baselines spanning naive retrieval, visual rendering, reranking, and context compression paradigms. Unless otherwise noted, all methods use Qwen3-VL-8B-Thinking as the generator backbone to ensure a fair comparison.

\textbf{Naive RAG}~\cite{lewis2020retrieval} directly concatenates all retrieved text documents with the query and feeds them to the generator. \textbf{Visual RAG} renders each retrieved document into an image using our rendering pipeline and passes all images to the generator at full resolution without any compression. \textbf{Static OCR Compress} applies a uniform compression ratio, equal to the average ratio produced by RAGOCR, to all rendered images, serving as a non-adaptive ablation of our method.

\textbf{Qwen3-Reranker-4B}~\cite{qwen3embedding} reranks the retrieved text passages by relevance score and feeds the reordered text to the generator. \textbf{Qwen3-VL-Reranker-2B}~\cite{qwen3vlembedding} operates on rendered images, reranking them before passing to the generator. These two methods remain top-k=5 documents.

\textbf{OSCAR}~\cite{louis2025oscar} is a query-dependent online soft compression method that jointly trains a compressor and generator; we directly evaluate its released checkpoints with Mistral-7B-Instruct-v0.2 and Mistral-Small-24B-Instruct-2501 as backbones, denoted as OSCAR-7B and OSCAR-24B respectively. \textbf{LongLLMLingua}~\cite{jiang-etal-2024-longllmlingua} performs token-level pruning guided by perplexity scores to accelerate long-context inference. \textbf{Provence}~\cite{chirkova2025provence} formulates context pruning as sequence labeling unified with reranking, dynamically adjusting the pruning ratio per query-context pair. \textbf{EXIT}~\cite{hwang2025exit} employs context-aware extractive sentence classification that adapts to query complexity. \textbf{FaviComp}~\cite{jung2024familiarity} generates abstractive compressions via ensemble decoding to lower perplexity with respect to the target model. For these four methods, we apply their respective compression strategies to the retrieved documents and then feed the compressed results to the shared Qwen3-VL-8B-Thinking~\cite{qwen3technicalreport} generator.

\subsection{Quantitative Results}

Table~\ref{table:performance of RAGOCR} presents the performance comparison across all five benchmarks. RAGOCR achieves 61.47\% average accuracy with only 2,284 RAG input tokens per case, attaining the highest scores on MedQA (69.21\%) and MMLU (68.87\%), the two most challenging reasoning-intensive benchmarks. This demonstrates that our query-aware optical compression effectively preserves the information most critical for complex multi-hop reasoning. Compared with Naive RAG (44.63\%), RAGOCR yields a substantial improvement of over 16 absolute points while reducing the average input from 17,914 tokens to 2,284 tokens, an approximately 87\% reduction.

Among methods that share the same generator backbone, RAGOCR outperforms all four text-based compression baselines. LongLLMLingua achieves only moderate compression with 8,813 tokens of RAG per case, yet obtains 53.70\% accuracy, indicating that token-level perplexity pruning struggles when the retrieved context is long and heterogeneous. EXIT retains the most tokens among compression methods, but reaches only 53.57\%, suggesting that extractive sentence selection alone is insufficient for complex retrieval scenario. Provence attains a reasonable balance with 3,036 tokens and 58.53\% accuracy, yet still falls behind RAGOCR by nearly 3 points despite consuming more tokens. FaviComp achieves aggressive compression with a compression ratio of about 16\%. However, it sacrifices accuracy substantially, highlighting the difficulty of maintaining task performance under extreme abstractive compression. OSCAR, which has the similar compression ratio as FaviComp, jointly trains compressor and generator on a dedicated backbone. Its larger version reaches 61.93\% with the least number of tokens. However, this comes at the cost of a substantially larger backbone, and a proprietary training pipeline. Its small version, with its smaller backbone, drops sharply to 45.17\%, underscoring the sensitivity of soft compression methods to backbone capacity.


A noteworthy observation is the significant gap between RAG and Visual RAG, which achieve 44.63\% and 57.16\% accuracy, respectively. Simply rendering text documents as images and feeding them at full resolution improves accuracy by over 12 points, empirically validating the higher information density of visual tokens over text tokens. Static OCR Compress further degrades from Visual RAG by 2 points to 55.02\%, confirming that uniform compression is suboptimal and that query-aware adaptation is essential—a gap that RAGOCR further widens, reaching 61.47\%.

\begin{table}[t]
\caption{Performance on SlideVQA, a benchmark with 20-page slide decks containing interleaved text and visual elements. All methods use Qwen3-VL-8B-Thinking as the generator. RAGOCR achieves the best results across all metrics.}
\label{tab:performance_ragocr_multimodal}
\renewcommand{\arraystretch}{0.95}
\setlength{\tabcolsep}{4pt}
\centering
\resizebox{\linewidth}{!}{
\begin{tabular}{lccc}
\toprule
\textbf{Metric} & \textbf{EM} & \textbf{F1-score} & \textbf{ANLS} \\
\midrule
Visual RAG             & 57.86 & 69.80 & 65.45 \\
Static OCR Compress    & 52.60 & 63.53 & 60.13 \\
Qwen3-VL-Reranker-2B   & 59.59 & 71.43 & 67.48 \\
\textbf{RAGOCR}        & 61.59 & 73.41 & 68.96 \\
\bottomrule
\end{tabular}}
\end{table}

\subsection{Multimodal Evaluation}

\begin{figure}[t]
  \includegraphics[width=\textwidth, trim=80 70bp 0 0, clip]{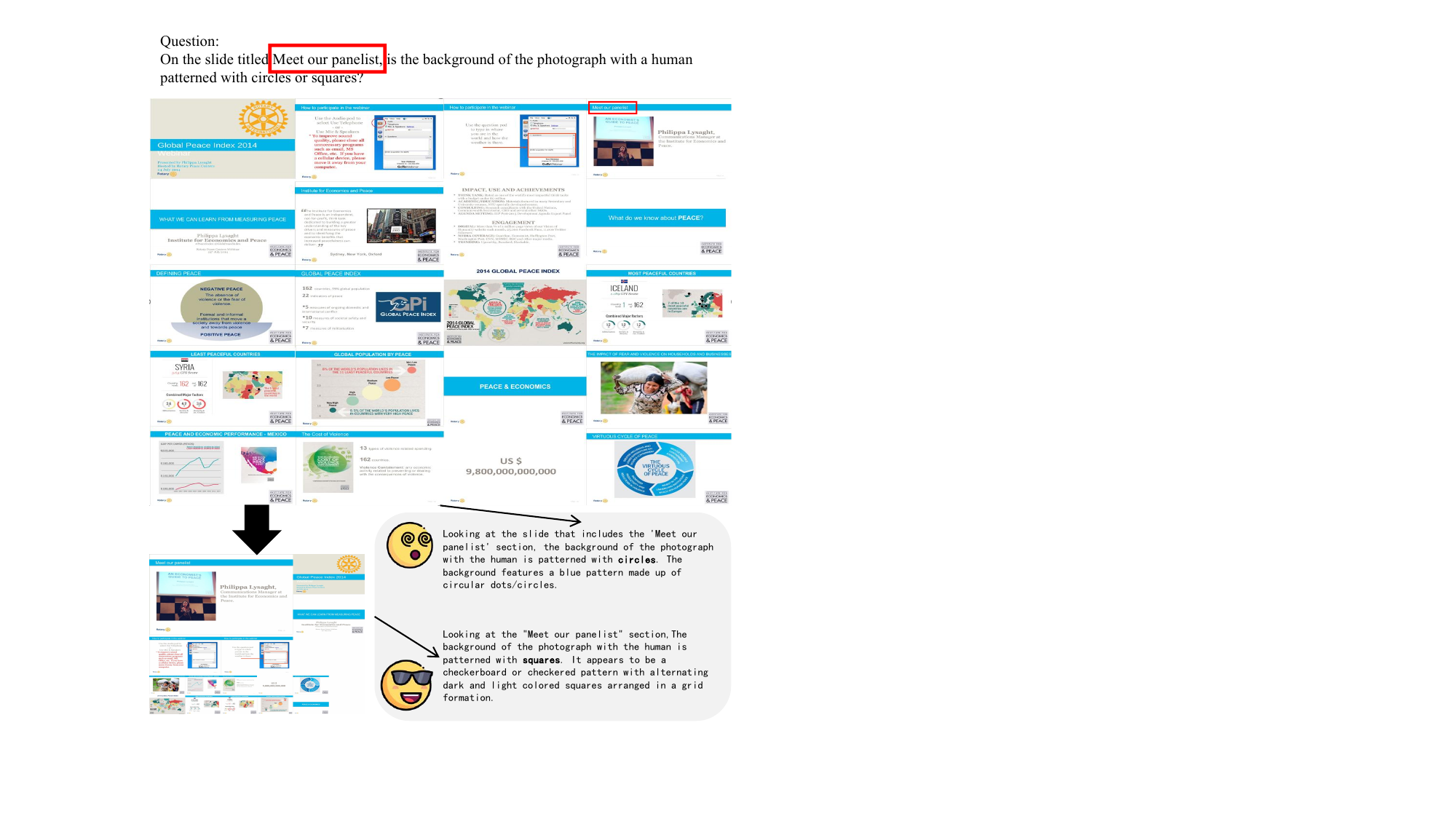}
  \caption{An example of data from SlideVQA, the key information is in image 4. RAGOCR precisely preserved the resolution of this image almost unchanged, while significantly compressing all other images.}
  \Description{An example of data from SlideVQA, the key information is in image 4. RAGOCR precisely preserved the resolution of this image almost unchanged, while significantly compressing all other images.}
  \Description{}
  \label{fig:case}
\end{figure}

To evaluate the generalizability of RAGOCR beyond pure-text documents, we conduct experiments on SlideVQA~\cite{tanaka2023slidevqa}, a multimodal question answering benchmark where each instance consists of a 20-page presentation slide deck containing interleaved text and visual elements. The 
ground-truth answer for each question is contained within one to three pages of the deck, challenging the model to locate relevant information across multiple slides and perform cross-page reasoning. Each ground-truth answer is a short free-form text span, and we prompt the generator to produce structured outputs from which we extract the predicted answer. We evaluate with three metrics: Exact Match (EM), which requires the prediction to be identical to the ground truth; F1-score, which measures token-level overlap between the prediction and the reference; and Average Normalized Levenshtein Similarity (ANLS), which computes the normalized edit distance between the two strings and is more tolerant of minor surface-level variations.
We train a compressor on the SlideVQA training set following the procedure described in Section~\ref{sec:compress}, and compare against three baselines: Visual RAG, which feeds all slide images at full resolution; Static OCR Compress, which applies the average compression ratio of RAGOCR uniformly to all slides; and Qwen3-VL-Reranker-2B, which reranks the 20 slides and retains the top-10 for generation.

As shown in Table~\ref{tab:performance_ragocr_multimodal}, RAGOCR achieves the best performance across all three metrics, reaching 61.59\% EM, 73.41\% F1, and 68.96\% ANLS. Compared with Visual RAG, RAGOCR improves EM by 3.73 points while processing substantially fewer visual tokens, demonstrating that adaptive compression can simultaneously reduce cost and improve accuracy by suppressing irrelevant slide content. Static OCR Compress degrades notably from Visual RAG, again confirming that uniform compression discards critical information from relevant pages. Qwen3-VL-Reranker provides a strong baseline by filtering out half of the slides, yet RAGOCR still surpasses it by 2.0 points in EM and 1.98 in F1, suggesting that continuous resolution control offers finer-grained relevance modulation than binary keep-or-discard selection. These results confirm that RAGOCR generalizes naturally to multimodal scenarios where documents contain rich visual elements beyond plain text. Figure ~\ref{fig:case} shows the effect of RAGOCR, which preserves important information and compresses the total tokens.

\subsection{Compressor with Reranker}

\begin{table}[t]
\caption{Exploring the role of RAGOCR in the RAG pipeline. The upper block shows different pipeline configurations: RAG, reranking with Qwen3-Reranker-4B, RAGOCR, and their combination (R\&R). The lower block evaluates RAGOCR as a standalone reranker by using its compression ratios as relevance scores, and Compress further applies optical compression to the filtered subset.}
\label{tab:reranker}
\renewcommand{\arraystretch}{0.9}
\setlength{\tabcolsep}{4pt}
\centering
\small
\begin{tabular}{lccccc}
\toprule
\textbf{Method} & \textbf{BioASQ} & \textbf{MedMCQA} & \textbf{MedQA} & \textbf{MMLU} & \textbf{PubMedQA} \\
\midrule
RAG        & 61.8 & 40.8 & 36.1 & 41.3 & 56.0 \\
Reranker         & 68.4 & 48.3 & 46.5 & 59.4 & 57.6 \\
RAGOCR  & 71.4 & 45.8 & 69.2 & 68.9 & 58.0 \\
R\&R  & 73.7 & 51.3 & 67.6 & 70.8 & 61.6 \\
\multicolumn{4}{l}{\small As Reranker} \\
filter         & 68.2 & 49.0 & 61.8 & 63.7 & 52.6 \\
compress  & 64.8 & 49.8 & 65.6 & 68.7 & 44.8 \\
\bottomrule
\end{tabular}
\end{table}

Beyond serving as a compression framework, we further explore the value of RAGOCR within the broader RAG pipeline. Specifically, we investigate two scenarios: using RAGOCR as a post-processing step after a dedicated reranker like Qwen-3-Reranker series, and replacing the reranker entirely with RAGOCR. Our exploration is motivated by the observation that RAGOCR learns to assign lower compression ratios to more query-relevant images and higher ratios to less relevant ones, so the compression ratios naturally encode a relevance ranking. We examine whether this emergent property can serve as an effective reranking signal, and how RAGOCR interacts with a dedicated reranker in the pipeline. Table~\ref{tab:reranker} summarizes the results.

We first examine whether RAGOCR and conventional reranking are complementary. R\&R (Reranker \& RAGOCR) first reranks the retrieved passages with Qwen3-Reranker-4B, then renders the reranked documents into images and processes them through RAGOCR for adaptive compression. This configuration achieves 63.60\%  weighted average accuracy, the highest among all configurations, surpassing both Qwen3-Reranker-4B alone (54.52\%) and RAGOCR alone (61.47\%). This indicates that text-based reranking and optical compression capture different aspects of document utility, and their combination yields additive gains. It also demonstrates that RAGOCR's optical compression maintains strong performance across varying numbers of input documents, whether operating on the full retrieval set or a smaller reranker-filtered subset.

We then investigate whether RAGOCR can entirely replace the dedicated reranker. We rank all retrieved documents by their compression ratios, where a lower ratio indicates higher relevance, and retain the same number of top-ranked passages as Qwen3-Reranker-4B. RAGOCR achieves a weighted average of 59.15\%, outperforming Qwen3-Reranker-4B (54.52\%) by nearly 5 points, and attaining comparable or superior results on four out of five benchmarks. Notably, Qwen3-Reranker series scores each document independently in a pointwise manner, whereas RAGOCR observes all documents jointly with the query in a single forward pass, effectively performing listwise ranking. This result suggests that the listwise ranking signal emerging from joint visual reasoning over rendered documents can be more effective than conventional pointwise text reranking, and that the render-then-rank paradigm may represent a promising direction for LLM-based rerankers in pure-text scenarios, fully leveraging the reasoning capabilities of VLMs.

Building upon the previous experiment, compress further applies optical compression to the retained documents, achieving 60.10\% weighted average. The improvement over Filter on benchmarks such as MedQA (61.8\% $\to$ 65.6\%) and MMLU (63.7\% $\to$ 68.7\%) indicates that adaptive resolution allocation continues to help the generator focus on the most informative regions even after irrelevant passages have been removed. However, Compress falls below R\&R (63.60\%), suggesting that the two-stage filter-then-compress pipeline starting from raw retrieval results loses some high-quality passages that Qwen3-Reranker-4B would have preserved. This confirms that RAGOCR and dedicated rerankers offer complementary strengths: the reranker excels at coarse-grained documents selection from noisy retrieval results, while RAGOCR provides fine-grained visual budget allocation within the selected set.

\subsection{Ablation on Image Shape}

\begin{table}[t]
\caption{Ablation on rendering shape. All configurations use font size $f{=}12$. W denotes the maximum canvas width, and H reports the resulting average canvas height across all rendered documents. Avg.\ is the weighted average accuracy across five benchmarks.}
\label{tab:shape}
\renewcommand{\arraystretch}{0.9}
\setlength{\tabcolsep}{4pt}
\centering
\small
\begin{tabular}{c|ccccc}
\toprule
\textbf{W/H} & \textbf{BioASQ} & \textbf{MedMCQA} & \textbf{MedQA} & \textbf{MMLU} & \textbf{PubMedQA} \\
\midrule
400/1703 & 58.31 & 51.50 & 66.77 & 68.23 & 56.40 \\
800/918  & 71.36 & 45.75 & 69.21 & 68.87 & 58.00 \\
1600/484  & 68.93 & 53.25 & 65.04 & 70.80 & 62.40 \\
\bottomrule
\end{tabular}
\end{table}

The rendering stage introduces two key hyperparameters: the maximum canvas width $w$ and the font size $f$. The former primarily controls the aspect ratio of the rendered image, while the latter affects the total number of visual tokens. We fix $f{=}12$ and vary $w$ across $\{400, 800, 1600\}$ to investigate how image shape influences downstream performance. As shown in Table~\ref{tab:shape}, when the canvas is narrow ($w{=}400$), each document is rendered as a tall image, resulting in the lowest accuracy of 59.85\%. Widening the canvas to $w{=}800$ produces a roughly balanced aspect ratio (800$\times$918) and improves performance to 61.47\%. Further increasing to $w{=}1600$ yields the best weighted average of 62.74\%, where each image becomes a wide, short strip. The trend suggests that horizontally laid-out text better aligns with how vision encoders process image patches, as wider canvases pack more characters per row and reduce the total number of rows, leading to more spatially compact and locally coherent visual representations. 
\section{Conclusion}

In this paper, we proposed RAGOCR, a framework that bridges retrieval-augmented generation and optical compression by rendering retrieved passages as images and adaptively assigning query-aware compression ratios through a GRPO-trained lightweight compressor. The compressor jointly examines all rendered images with the input query and allocates higher visual resolution to more relevant passages while aggressively down-scaling less pertinent ones, achieving substantial token reduction without sacrificing downstream answer quality. Extensive experiments on multiple knowledge-intensive QA benchmarks demonstrate that RAGOCR consistently outperforms existing context compression methods in both accuracy and efficiency. Beyond compression, we discovered that the learned compression ratios inherently encode passage relevance, enabling the compressor to double as an effective reranker that surpasses dedicated reranking models, suggesting that optical compression and relevance estimation can be unified within a single lightweight model. We further validated the generalizability of RAGOCR on a multimodal benchmark involving rich images and text, confirming its applicability beyond pure-text scenarios. We hope this work inspires future exploration of a unified retrieve-compress-rank paradigm for retrieval-augmented generation.
\bibliographystyle{ACM-Reference-Format}
\bibliography{sample-base}

\end{document}